\documentclass[twocol]{ametsocV6.1}
\bibpunct{(}{)}{;}{a}{}{,}

\usepackage{color}

\title{Comparing Explanation Methods for Traditional Machine Learning Models Part 2:  Quantifying Model Explainability Faithfulness and Improvements with Dimensionality Reduction}

\authors{Montgomery L. Flora\aff{a,b,e}\correspondingauthor{Montgomery Flora, monte.flora@noaa.gov}, 
Corey K. Potvin,\aff{b,c,e}, 
Amy McGovern\aff{c,d,e}
Shawn Handler\aff{a,b,*}\thanks{*Current Affiliation: Verisk Analytics Incorporated, Jersey City, New Jersey},
}

\affiliation{\aff{a}{Cooperative Institute for Severe and High-Impact Weather Research and Operations, University of Oklahoma, Norman, Oklahoma}\\
\aff{b}{NOAA/OAR/National Severe Storms Laboratory, Norman, Oklahoma}\\
\aff{c}{School of Meteorology, University of Oklahoma, Norman, Oklahoma}\\
\aff{d}{School of Computer Science, University of Oklahoma, Norman, Oklahoma}\\
\aff{e}{NSF AI Institute for Research on Trustworthy AI in Weather, Climate, and Coastal Oceanography}\\
}

\abstract{Machine learning (ML) models are becoming increasingly common in the atmospheric science community with a wide range of applications. To enable users to understand what an ML model has learned, ML explainability has become a field of active research. In Part I of this two-part study, we described several explainability methods and demonstrated that feature rankings from different methods can substantially disagree with each other. It is unclear, though, whether the disagreement is overinflated due to some methods being less faithful in assigning importance. Herein, "faithfulness" or "fidelity" refer to the correspondence between the assigned feature importance and the contribution of the feature to model performance. In the present study, we evaluate the faithfulness of feature ranking methods using multiple methods. Given the sensitivity of explanation methods to feature correlations, we also quantify how much explainability faithfulness improves after correlated features are limited. Before dimensionality reduction, the feature relevance methods [e.g., SHAP, LIME, ALE variance, and logistic regression (LR) coefficients] were generally more faithful than the permutation importance methods due to the negative impact of correlated features. Once correlated features were reduced, traditional permutation importance became the most faithful method. In addition, the ranking uncertainty (i.e., the spread in rank assigned to a feature by the different ranking methods) was reduced by a factor of 2-10, and excluding less faithful feature ranking methods reduces it further. This study is one of the first to quantify the improvement in explainability from limiting correlated features and knowing the relative fidelity of different explainability methods.
}

\begin{document}
\maketitle


\statement{Complex machine learning models require outside methods to help understand how they work. However, as found in the first part of this two-part study, explainability methods frequently disagree on the ranking of the most important features, making it difficult to know which features are truly the most important. In the present study, our objective is to assess the faithfulness of different feature ranking methods. We find that different feature ranking methods indeed have different degrees of faithfulness, but generally provide plausible rankings, which could lead to misunderstanding the model’s behavior. On the other hand we also find that  reducing correlated features and excluding less faithful methods enables better explanations of model behavior. }

\section{Introduction}\label{intro}
Many ML models are highly nonlinear and high-dimensional making them difficult to understand without the use of external methods \citep{McGovern+etal2019_blackbox}. As discussed in \citet{PartI} (hereafter Part I), we can use post hoc techniques, known as explainability methods (defined in Part I), to gain some understanding of how these complex models work. However, in agreement with previous studies (e.g., \citealt{McGovern+etal2019_blackbox, Satyapriya+etal2022_disagree}), we demonstrated in Part I that explanation methods can disagree with each other. Although some disagreement is justifiable since different explanation methods are designed for different tasks (e.g., feature relevance versus feature importance; see Section 4a.1 of Part I), it is unclear whether disagreement is inflated by less faithful methods. Herein, "faithfulness" or "fidelity" refer to the correspondence between the assigned feature importance and the contribution of the feature to model performance. For example, permutation importance may misassign individual importance when features are correlated (e.g., \citealt{Strobl+etal2007, Strobl+etal2008, Nicodemus+etal2010, Gregorutti+etal2015_grouped, Gregorutti+etal2017_corr}), and thus produce less faithful feature rankings in many meteorological applications. However, little research has been done to compare the faithfulness of a wide range of feature ranking methods, especially in the context of atmospheric science-relevant datasets. \citet{Covert+etal2020} is likely the first study to compare the faithfulness of multiple feature ranking methods, but their datasets were not atmospheric science-based and did not include feature ranking methods commonly used in atmospheric sciences (e.g., multi-pass permutation importance; \citealt{McGovern+etal2019_blackbox, Jergensen+etal2020, Handler+etal2020, Lagerquist+etal2021}). 

In this study, we explore the faithfulness of the feature ranking methods introduced in Part I using approaches introduced in \citet{Covert+etal2020} and evaluate whether the ranking disagreement can be reduced by identifying and excluding lower-fidelity methods. Since some of the explainability methods explored in this study are sensitive to correlated features, we also evaluate the impact of dimensionality reduction on the ranking results and on the fidelity of individual methods. These efforts are important, since there is a \textit{disagreement problem} for explainability methods (see Part I) where ML practitioners find it difficult to choose among explainability methods that provide different feature rankings and are forced to rely on heuristics to make decisions \citep{Satyapriya+etal2022_disagree}. These disagreements can interact with confirmation bias and cause practitioners to make incorrect assumptions about the model behavior. 

As in Part I, we use ML models developed for severe weather prediction \citep{Flora+etal2021} and sub-freezing road surface temperature prediction \citep{Handler+etal2020}. Full descriptions of each dataset are provided in Part I and therefore are omitted from this paper. We limit our focus to the logistic regression models from \citet{Flora+etal2021} and the random forest trained in \citet{Handler+etal2020}. Using both datasets, we outline pertinent discussions on ML explainability and to begin to generalize the behavior of different feature ranking methods. A full description of the explainability methods used in this study are provided in Part I and is omitted from this paper. All the methods demonstrated here are available in the Python package scikit-explain \citep{Flora+Handler}, which was developed by the authors.

The structure of this paper is as follows. Section \ref{sec:reduce_model_complexity} describes our methods for reducing model complexity and the metrics we used to quantify model complexity.  We present the results in Section ~\ref{sec:results} while the conclusions and limitations of the study are discussed in Section ~\ref{sec:conclusions}. 


\section{Data and Machine Learning Methods} \label{method} \label{sec:log_reg}
To generalize the results, this study uses two different types of meteorological datasets: severe weather prediction and sub-freezing road surface prediction. The severe weather dataset is derived from Warn-on-Forecast System (WoFS) output using an object-based approach while the road surface dataset is a grid-based nowcasting dataset based on hourly output from the High-Resolution Rapid Refresh (HRRR). More details on the datasets are provided in Part I, \citet{Handler+etal2020}, and \citet{Flora+etal2021}.

For this study, we used classification random forests and logistic regression models available in the Python sci-kit learn package \citep{Pedregosa+etal2011}. Consistent with \citet{Handler+etal2020} and \citet{Flora+etal2021}, the random forest is applied to the road surface dataset to predict whether a road will freeze while logistic regression is applied to the WoFS dataset to predict whether a storm track will be associated with a severe hail, severe wind, or tornado report, respectively. The hyperparameter optimization is described in the the appendix. 

\section{Reducing Model Complexity and Limiting Correlated Features}\label{sec:reduce_model_complexity}
\subsection{Rashomon Sets and Multi-Objective Optimization} 
Before discussing the model explainability, we address the fact that disparate ML models can fit the data equally well, but rely on different covariate information. This phenomenon is known as the \textit{Rashomon effect} \citep{Breiman2001, Fisher+etal2018, Rudin+etal2018, Rudin+etal2021} and occurs when there are multiple valid descriptions of the same event. \citet{Fisher+etal2018} formalized the idea of the \textit{Rashomon set} for a given dataset as the set of models that have expected accuracy within some defined error of the best possible model. This set can be thought of as including all models that might be obtained depending on the model developer's choices regarding data measurement, pre-processing, filtering, model class and parameterization, and so on.  Although ML development is often concerned with how well the model generalizes to unseen data \citep{Breiman2001}, less attention is paid to producing an explainable model. In line with \citet{Molnar+etal2019}, we argue that ML model development should be treated like a multi-objective optimization problem where the goal is not solely to develop the best-performing model, but to train simpler, well-performing models (ones that are within the Rashomon set) that are therefore more explainable.
Often the literature presents the following dichotomy: decreasing (increasing) model complexity reduces (improves) model performance but improves (limits) explainability and interpretability. However, for some datasets, models can be simplified with little or no loss of skill (e.g., \citealt{Hand_2006, Kuhle+etal2018, Rudin+etal2018, Rudin+etal2021, Christodoulou+etal2019, Chakraborty+etal2021, Flora+etal2021, Molnar+etal2021}). If the Rashomon set is sufficiently large, a simpler, well-performing model is more likely to exist \citep{semenova2021}. Therefore, our goal is not only to produce the best score on an independent testing dataset, but also to improve model explainability by limiting the number of features, spurious feature interactions, or the complexity of feature effects \citep{Molnar+etal2019}. This philosophy is used in the following section when dimensionality reduction is applied to the datasets. 

\subsection{Feature Selection and Reducing Multicollinearity}\label{sec:removing_corr}
One approach to reducing model complexity and improving model explainability is removing redundant features.  While strong feature multicollinearity may not adversely affect model performance, parsing feature-unique model contributions becomes difficult. For example, for regression-based learning algorithms, if two features are highly correlated (e.g., $\rho \geq 0.9$ ) their regression coefficients can spuriously have opposite signs, which does not impact model performance (due to compensatory effects), but it misrepresents the relationship of the affected feature to the target variable.

There are other reasons to reduce information redundancy.  First, decreasing the number of features reduces the overall computation time. For example, for the real-time prediction of storm longevity, \citet{McGovern+etal2019_stormlong} found that reducing the feature set decreased pre-processing time while having little impact on model performance. Second, shrinking the feature set can decrease the likelihood of overfitting by reducing opportunities for ML models to learn noise or spurious feature interactions. Third, explanations of model behavior become more compact as we reduce the set of features \citep{Molnar+etal2019}.  

\begin{figure*}[t]
    \noindent\includegraphics[width=38pc,angle=0]{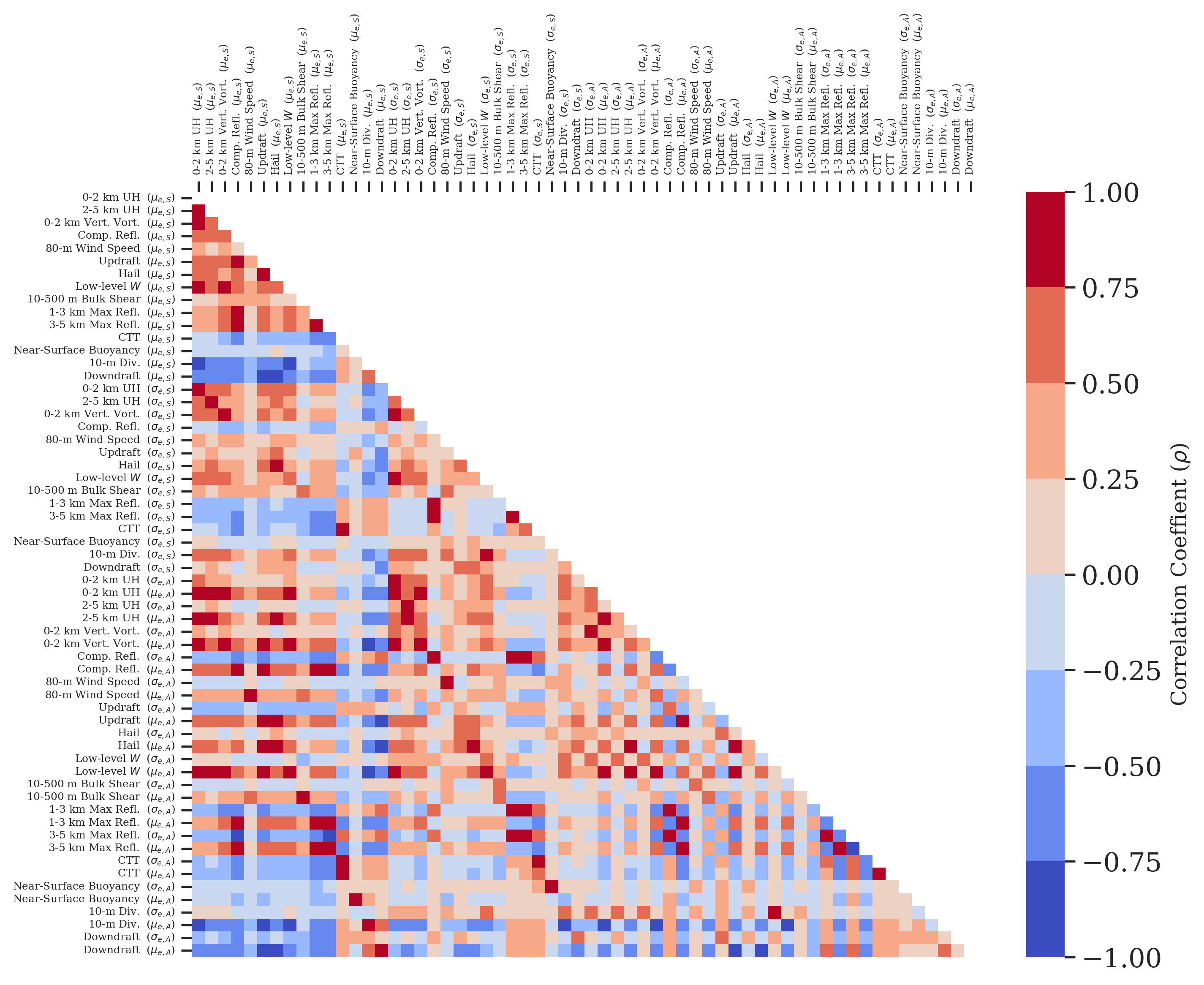}
    \caption{Linear correlation coefficient matrix for the spatial ($S$) and amplitude ($A$) intra-storm features in the WoFS dataset. Ensemble mean and standard deviation are indicated by $\mu_e$ and $\sigma_e$, respectively. Additional details are provided in \citet{Flora+etal2021}. }
     \label{fig:wofs_correlation_matrix}
\end{figure*}

    The datasets in this study are reduced by a combination of manual feature removal and a feature selection method based on logistic regression with L1 regularization. The feature selection method trains a logistic regression model with L1 regularization, which incentivizes the model to remove irrelevant features, especially multicollinear features thereby improving model sparsity. For example, if two features are nearly identical, then L1 regularization will set the coefficient of one of them to zero. Unlike sequential selection methods, the L1-norm-based method is computationally feasible for large datasets but requires a linear model to perform well on the data (which was true for both datasets). 

\begin{figure}[t]
    \noindent\includegraphics[width=20pc,angle=0]{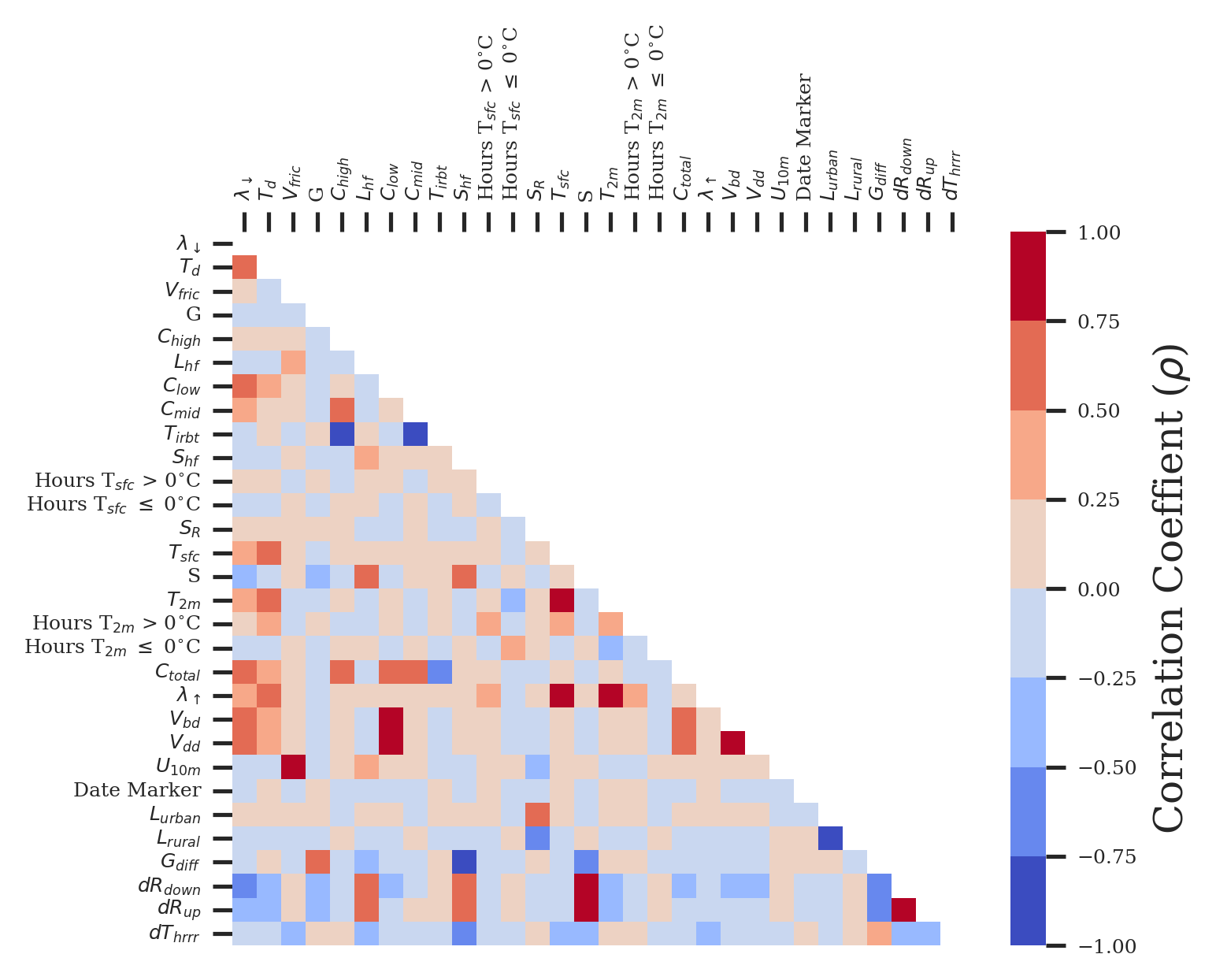}
    \caption{Linear correlation coefficient matrix for the sub-freezing road surface temperature dataset. Description of the features provided in \citet{Handler+etal2020}. }
     \label{fig:road_surface_correlation_matrix}
\end{figure}

    Our first step in the feature reduction process was to compute a linear correlation coefficient matrix for both datasets. For the WoFS dataset, we found that the spatial average intrastorm features were often highly correlated with the amplitude intrastorm features ($\rho \geq 0.7$; Fig.~\ref{fig:wofs_correlation_matrix}) and the removal of spatial-based intra-storm features did not adversely affect model performance. For the road surface dataset (Fig.~\ref{fig:road_surface_correlation_matrix}), although there were fewer strong correlations, we limited the feature set to near-surface features and removed many radiative features given their correlation with temperature variables. After manually removing these features from the two datasets, we applied L1 regularization, whose strength is governed by the parameter $C$.  We found that $C=0.0075$ plus a cutoff threshold of $10^{-5}$ (below which a coefficient is considered zero) was sufficient for reducing the feature set. For the three severe weather hazards, approximately 30 or fewer features (of the original 113 features) were retained after the manual and L1-norm-based feature selection. The road surface dataset was reduced from 30 to 11 features. As will be discussed in Section \ref{sec:results}\ref{sec:results_feature_rankings}, the final number of features for each of the four data sets is consistent with the number of features that capture most of the original model performance. For the four datasets, the final sets of features are shown in Tables \ref{table:reduce_features_tornado}-\ref{table:reduce_features_road}.

    \begin{table*}[h]
    \caption{Reduced feature set for the tornado model. $\mu_e$ and $\sigma_e$ refer to the mean and spread of the ensemble, respectively. $A$ and $S$ refer to whether the feature is amplitude- or spatially-based, respectively. } \label{table:reduce_features_tornado}
    \begin{center}
    \begin{tabular}{llccc}
    \hline \hline

Storm &  Environment &   Morphology \\
\hline
2-5 km Updraft Helicity ($\mu_{e,A}$) & ML CAPE  ($\mu_{e,S}$) & Major axis length \\
0-2 km Avg. Vertical Vorticity ($\mu_{e,A}$) & 0-6 km V Shear  ($\mu_{e,S}$) & Minor axis length \\
Composite Reflectivity  ($\mu_{e,A}$) & 0-1 km V Shear  ($\mu_{e,S}$) &  \\
Column-maximum Updraft  ($\mu_{e,A}$) & ML LCL  ($\mu_{e,S}$) \\
Hail  ($\mu_{e,A}$) & 500 mb Geopotential height  ($\mu_{e,S}$) \\
                   & ML CIN  ($\sigma_{e,S}$) \\

\hline
\end{tabular}
\end{center}
\end{table*}

\begin{table*}[h]
\caption{As in Table \ref{table:reduce_features_tornado}, but for severe hail.  } \label{table:reduce_features_hail}
\begin{center}
\begin{tabular}{llccc}
\hline \hline
Storm &  Environment &    Morphology/Other \\
\hline
2-5 km Updraft Helicity ($\sigma_{e,A}$) & 0-3 km Storm Relative Helicity  ($\mu_{e,S}$) & Major axis length \\
2-5 km Updraft Helicity ($\mu_{e,A}$) & ML CIN  ($\mu_{e,S}$) & Minor axis length \\
Composite Reflectivity  ($\sigma_{e,A}$) & 0-6 km U Shear  ($\mu_{e,S}$) \\
Composite Reflectivity ($\mu_{e,A}$) & 0-1 km U Shear  ($\mu_{e,S}$) \\
80-m Wind Speed  ($\sigma_{e,A}$) & 10-m U  ($\mu_{e,S}$) \\
Updraft ($\sigma_{e,A}$) & 10-m V  ($\mu_{e,S}$) \\
Updraft ($\mu_{e,A}$) & 0-3 km Lapse Rate  ($\mu_{e,S}$) \\ 
0-1 km Updraft ($\sigma_{e,A}$) & T$_{500}$  ($\mu_{e,S}$) \\ 
10-500 m Bulk Shear  ($\sigma_{e,A}$) & 700 mb Geopotential Height  ($\mu_{e,S}$) \\
Cloud Top Temperature  ($\sigma_{e,A}$) & 500 mb Geopotential Height ($\mu_{e,S}$) \\
Cloud Top Temperature  ($\mu_{e,A}$) & T$_{d, 500}$  ($\mu_{e,S}$) \\
Downdraft  ($\mu_{e,A}$) &  ML CAPE  ($\sigma_{e,S}$) \\
Near-surface $\theta_e$ deficit ($\sigma_{e,A}$) & ML CIN  ($\sigma_{e,S}$) \\ 
Near-surface $\theta_e$ deficit ($\mu_{e,A}$) & 500-700 mb Lapse Rate  ($\sigma_{e,S}$) \\
                                              & 500 mb Geopotential Height ($\sigma_{e,S}$) & \\

\hline
\end{tabular}
\end{center}
\end{table*}

\begin{table*}[h]
\caption{As in Table \ref{table:reduce_features_tornado}, but for severe wind. } \label{table:reduce_features_wind}
\begin{center}
\begin{tabular}{llccc}
\hline \hline
Storm &  Environment &    Morphology/Other \\
\hline

0-2 km Avg. Vertical Vorticity ($\mu_{e,A}$) & 0-1 km Storm Relative Helicity  ($\mu_{e,S}$) & Major axis length \\
Composite Reflectivity  ($\mu_{e,A}$) & 0-6 km U Shear  ($\mu_{e,S}$) & Minor axis length \\
80-m Wind Speed  ($\mu_{e,A}$) & 0-1 km U Shear  ($\mu_{e,S}$) \\
Hail  ($\sigma_{e,A}$) & ML LCL  ($\mu_{e,S}$) \\
Hail  ($\mu_{e,A}$) &  10-m U  ($\mu_{e,S}$) \\
0-1 km Updraft  ($\mu_{e,A}$) & 500-700 mb Lapse Rate  ($\mu_{e,S}$) \\
10-500 m Bulk Shear  ($\mu_{e,A}$) & 0-3 km Lapse Rate  ($\mu_{e,S}$) \\
Near-surface $\theta_e$ deficit  ($\sigma_{e,A}$) & T$_{d, 850}$  ($\mu_{e,S}$) \\
                                                  & 0-6 km U Shear  ($\sigma_{e,S}$) \\
                                                  & 10-m U  ($\sigma_{e,S}$) \\
                                                  & 10-m U  ($\sigma_{e,S}$) \\
                                                  & 850 mb Geopotential Height  ($\sigma_{e,S}$) \\
                                                  & 500 mb Geopotential Height ($\sigma_{e,S}$) \\
                                                  & T$_{d, 500}$  ($\sigma_{e,S}$) \\

\hline
\end{tabular}
\end{center}
\end{table*}

\begin{table*}[h]
\caption{Reduced feature set for the road surface model.} \label{table:reduce_features_road}
\begin{center}
\begin{tabular}{llccc}
\hline \hline
Temperature-based & Radiation-Based & Miscellaneous \\ 
\hline

Surface Temperature ($T_{sfc}$) & Downward longwave radiation flux ($\lambda_{\downarrow}$) & Absolute difference between current date and 10 Jan (Date Marker) \\
Hours $T_{sfc}$ $<= $0$\circ$C & Simulated Brightness Temperature ($T_{irbt}$) & Low cloud cover percentage ($C_{low}$) \\
2-m Dewpoint Temperature & Incoming shortwave radiation ($S$) & \\
                    & Ground Flux ($G$) &  \\ 
                    & Surface Latent heat flux ($L_{hf}$) & \\ 
                    & Surface Sensible heat flux ($S_{hf}$) & \\

\hline
\end{tabular}
\end{center}
\end{table*}

As discussed above, our goal is not to define the most skillful model, but to produce a feature subset that does not substantially harm the model performance while greatly improving explainability. By increasing the model sparsity and reducing multicollinearity, we gain a better holistic view of how variables interact jointly rather than individually with the target variable and a more compact explanation of the relationships learned.  

\subsection{Quantifying Model Complexity}\label{sec:ale_complex}
Recently, \citet{Molnar+etal2019} introduced two metrics, the interaction strength statistic (IAS) and main effect complexity (MEC), for quantifying ML model complexity using the accumulated local effects (ALE; \citealt{Apley+etal2016}). The ALE for the feature $x_j$ is :
\begin{equation}
    ALE_j(x_j) = \int_{\min(z_j)}^{x_j} \mathbb{E} \Bigg[\frac{\partial f(\mathbf{X})}{\partial X_j} \Big| X_j = z_j \Bigg] dz_j - c
\end{equation}
where $f$ is the ML model, $\mathbf{X}$ is the set of all features, $z_j$ are the values of $x_j$, c is the integration constant (set equal to the average $ALE_j (x_j)$ so the effect is centered). To compute the feature effect, ALE calculates the expected change in prediction over different conditional distributions and then integrates them (cumulative sum).  By computing the effect of the feature from conditional rather than marginal distributions, ALE isolates the effect from the effects of all other features. More details on the ALE computation are provided in \citet{InterpretMLTextbook}. 

From \citet{Molnar+etal2019}, any high-dimensional prediction function (e.g., an ML model) can be decomposed into a sum of components with increasing dimensionality:
\begin{equation}\label{eqn:ale_decompose}
    f(x) = \overbrace{f_o}^{\text{Intercept}} + \overbrace{\sum_{j=1}^{P} f_j (x_j)}^{\text{1st order effects}} +  \overbrace{\sum_{j<k}^{P} f_{jk} (x_j,x_k)}^{\text{2nd order effects}} + ... + \overbrace{f_{1,...,P}(x_{1,...,}x_P}^{\text{P-th order effects}}),
\end{equation}
where $P$ is the number of features. Using Equation \ref{eqn:ale_decompose}, we can approximate an ML model by using the average model prediction for $f_o$, the first-order ALE, and all second-order or higher effects [denoted by $I(x)$]: 
\begin{equation}
    f(x) = \frac{1}{N}\sum_{i=1}^N f(x^{(i)}) + \sum_{j=1}^{P} ALE_j (x_j) + I(x),
\end{equation}
where $x^{(i)}$ is the $i$th training example. We can define the IAS as an approximation error of the first-order effects with respect to the original model predictions:
\begin{equation}\label{eqn:ias}
    IAS = \frac{\sum_{i=1}^N \Big( f[x^{(i)}] - f_{1stOrder}[x^{(i)}]\Big)^2}{\sum_{i=1}^N \Big( f[x^{(i)}] - f_0\Big)^2},
\end{equation}
where $f_{1stOrder} = f_0 + ALE_1(x_1) +...+ALE_P(x_P)$ ($f_o$ is defined as the average model prediction). If IAS = 0, then a ML model is perfectly approximated by first-order effects and has no feature interactions. 

To describe the shape of the first-order ALE curves, \citet{Molnar+etal2019} introduced the MEC statistic:
\begin{equation}\label{eqn:mec}
    \text{MEC} = \frac{1}{\sum_{j=1}^P \sigma^2_{ALE_j(x_j)}} \sum_{j=1}^{P} \sigma^2_{ALE_j(x_j)} MEC_j
\end{equation}
where $MEC_j$ is the number of line segments needed to approximate the curve $ALE_j(x_j)$ with a piecewise linear function and $\sigma^2_{ALE_j(x_j)}$ is the variance of $ALE_j(x_j)$. The MEC algorithm starts by approximating the ALE curve with a linear model and measuring if the approximation is within tolerance (i.e., $R^2 > 1 - \epsilon$ where $\epsilon$ is 0.05 for this study). If not, the domain is repeatedly split into two parts based on each value of $x_j$, and each part is approximated with a linear model. If the tolerance is not met, the best split is maintained and then each subdomain is divided. Until the tolerance is met, the domain is divided into more parts (while maintaining the former best splits). For a fully linear model the MEC should equal 1 while higher values indicate more nonlinear first-order effects. 

Using the IAS and MEC, we quantify the reduction in the ML models' complexity with the reduced feature sets. Ideally, the more a model is simplified, the more its explainability increases. To improve the IAS and MEC estimates, ALE curves were calculated from bootstrapped training sets ($N$ = 100) for the original and reduced feature set and the mean IAS and MEC were calculated. 

\section{Results}\label{sec:results}

\begin{figure}[t]
    \noindent\includegraphics[width=20pc,angle=0]{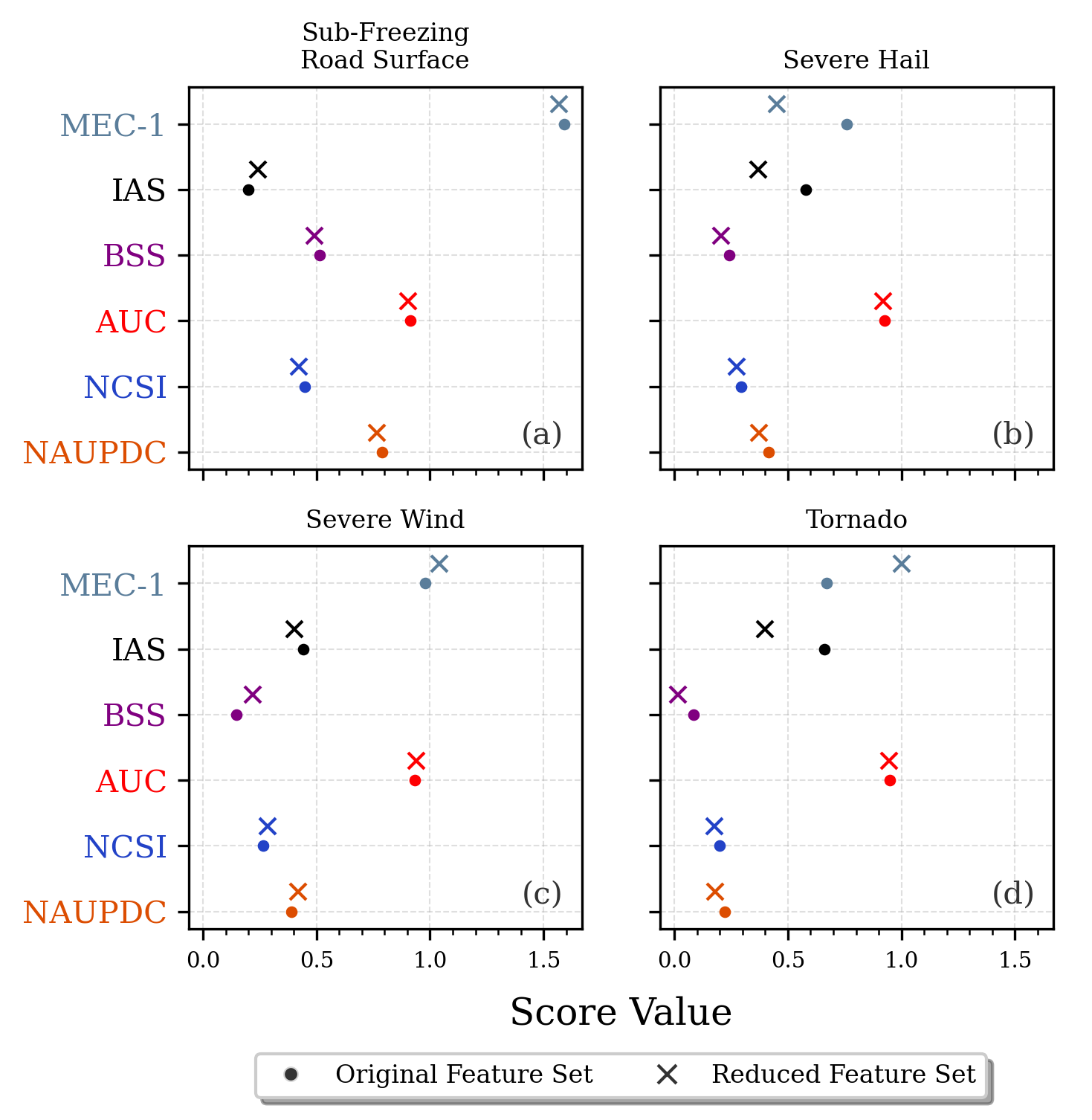}
    \caption{Comparison of the bootstrap average (N=1000) NAUPDC (orange), NCSI (blue), AUC (red), and BSS (purple) for the (a) sub-freezing road surface, (b) severe hail, (c) severe wind, (d) tornado models on their respective testing datasets. To quantify the change in model complexity, the interaction strength (IAS; black) and main effect complexity (MEC; bluish gray) from \citep{Molnar+etal2019} are also shown for each hazard (computed on the training dataset). MEC-1 is shown so that all statistics can be shown on a similar x-axis. }
     \label{fig:compare_scores}
\end{figure}

\subsection{Comparing model performance of original feature set versus reduced feature set} 
Figure \ref{fig:compare_scores} compares the model performance and complexity statistics of the original and reduced feature set for all four datasets. The area under the receiver operating characteristic curve (AUC; \citealt{Metz1978}) and Brier Skill Score (BSS;\citealt{Hsu+Murphy1986}) are common verification metrics for probabilistic predictions, while NAUPDC and the normalized critical success index (NCSI; \citealt{Flora+etal2021, Miller+etal2021}) have recently been introduced for the prediction of rare events. The ML models trained on the reduced feature sets produced performance practically similar to those trained on the original feature sets. For the severe hail and tornado models (Fig.~\ref{fig:compare_scores}b,d), reducing the feature set resulted in an IAS (Eqn. \ref{eqn:ias}) drop of approximately 0.2, while for the models of severe wind and subfreezing road temperatures (Fig.~\ref{fig:compare_scores}a,c), the IAS was relatively unchanged. We can interpret IAS as the percentage of the model prediction explained by higher-order effects, so a decrease from 0.5 to 0.3 indicates a 40$\%$ decrease in the contribution of higher-order effects to the model prediction. When we remove less important features, we might anticipate a lower IAS, as the ML model is less prone to creating spurious feature interactions by fitting noise in the training data. The drop in IAS for tornado and severe hail models coupled with similar model performance despite the feature set reduction indicates that these models are relying more on first-order effects and that some feature interactions in the original dataset were likely spurious. As for MEC (eqn. \ref{eqn:mec}), a simple measure of the non-linearity of first-order effects, the impact of removing features is inconsistent. We found that in the original dataset, there were many features that had small but non-negligible ALE variance (i.e., nearly flat curves), which biased the MEC lower. This may partly explain why the MEC increased when the tornado feature set was reduced. It is possible that the first-order effects became stronger in the reduced feature set (e.g., increasing the ALE variance) and therefore increased the MEC. The MEC values between 1.5 and 2.5 indicate that the first-order effects are fairly simple in all four models, which helps to summarize the model behavior. 

\subsection{Assessing the Validity of Feature Ranking Methods}\label{sec:results_feature_rankings}
\begin{figure*}[t]
    \noindent\includegraphics[width=38pc,angle=0]{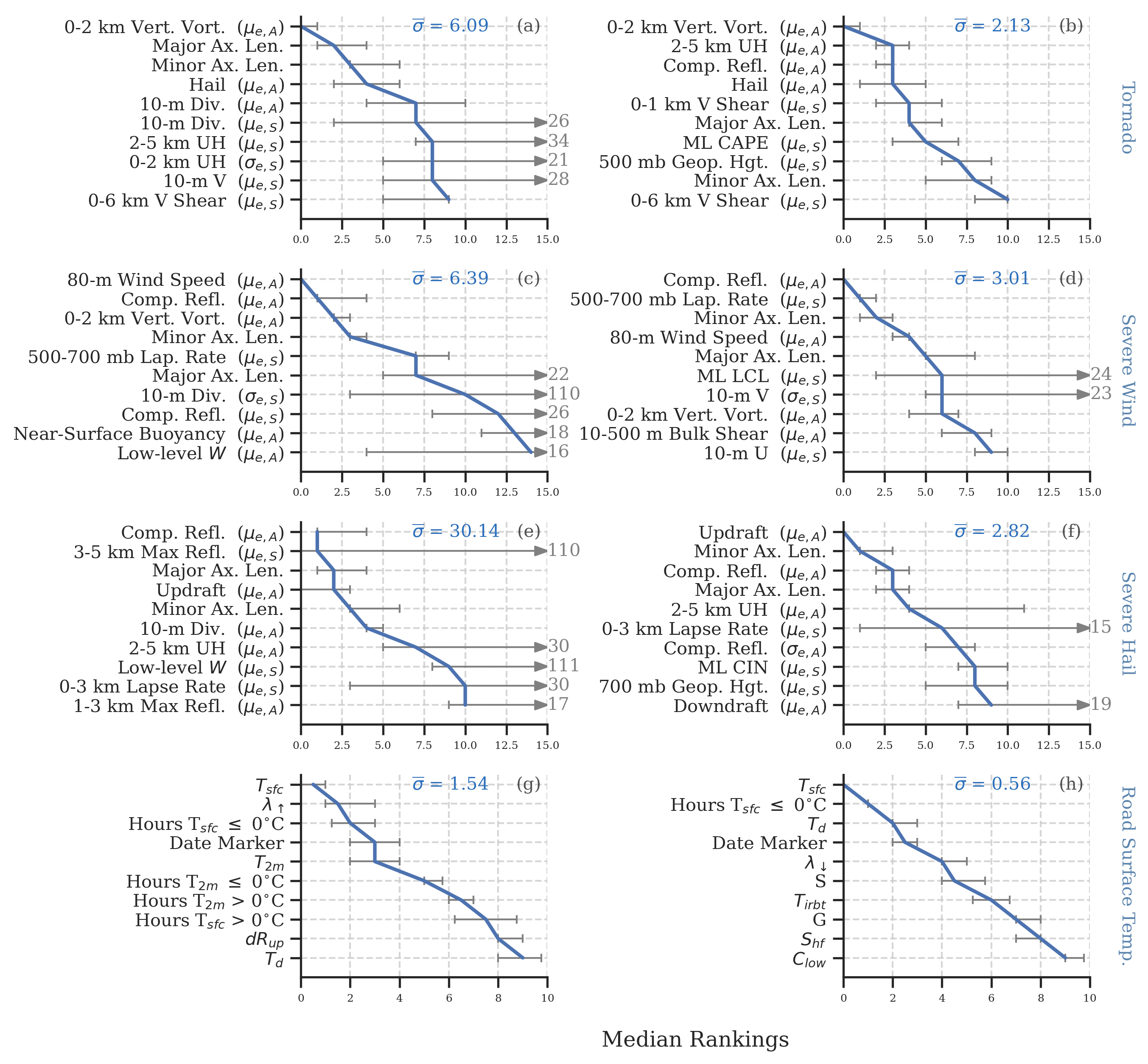}
    \caption{Median feature ranking for the original (left column) and reduced (right column) feature datasets for the tornado (first row), severe wind (second row), severe hail (third row), and road surface temperature (last row) models. Uncertainty (gray bars) in ranking shown with interquartile percentile range. In cases of very large uncertainties, the upper bound truncated at the edge of the plot and the value is provided. In the upper right of each panel, the weighted average feature ranking uncertainty statistic ($\overline{\sigma}$) is provided. $\mu_e$ and $\sigma_e$ indicate ensemble mean and standard deviation, respectively, while the subscripts $A$ and $S$ indicate amplitude or spatial statistics, respectively. }
     \label{fig:median_rankings}
\end{figure*}

Fig.~\ref{fig:median_rankings} shows the median feature ranking for the original and reduced feature sets. Despite the fact that the original and reduced feature models have similar performance (Fig.~\ref{fig:compare_scores}) there is only partial agreement among the ranking methods on the top 3-5 features and their respective ranks. The noticeable difference in the rankings between the original and reduced models illustrates that feature ranking methods are designed to explain the model rather than the data, and for our datasets, the Rashomon set contains models with a wide range of model reliance on a given feature \citep{Fisher+etal2018}. For a given dataset, the top 3-5 features from the original or reduced models are plausible sets of important features, but either leads to a different conclusion about the dataset. For example, the original tornado and severe wind datasets, the main features are largely based on spatial low-level features (e.g., 0-2 km vertical vorticity, 10-m divergence, and 80-m wind speed; Fig.~\ref{fig:median_rankings}a, c) while the reduced feature sets are based on amplitude-based mid-level features [e.g., hail diameter, 2-5 km UH, mid-level lapse rate, and composite reflectivity; Fig.~\ref{fig:median_rankings}b,d).  When using ML for knowledge discovery, it is crucial to keep this limitation in mind. Trusting any plausible explanation will lead to confirmation bias \citep{McGovern+etal2019_blackbox, Molnar+etal2021, Ghassemi+etal2021}. 

For the original severe wind and severe hail datasets (Fig.~\ref{fig:median_rankings}c,e), there are rather large rank uncertainties. For example, 3-5 km maximum reflectivity (3-5 km Max Refl) has a median ranking of 1 (tied for the top most important feature) for the severe hail model while some methods (e.g., the forward permutation methods) have it ranked as one of the worst features. We showed in Part I that the logistic regression had learned the opposite effect for 3-5 km Max Refl, which is due to its strong correlation with features that are more predictive of large hail (e.g., HAILCAST-predicted hail size, updraft speed, composite reflectivity). If 3-5 km Max Refl was removed from the original model, but the features it is correlated with were not, then the performance of the model would suffer due to the loss of the compensating effect, and 3-5 km Max Refl would be assigned high feature importance. However, if 3-5 km Max Refl was retained but its correlated features were removed, its compensating effect within the model would not be useful, rendering it unimportant. This result highlights that the backward and forward permutation importance methods are measuring different aspects of feature importance. Furthermore, 3-5 Max Refl was removed from the original severe hail feature set (Table ~\ref{table:reduce_features_hail}) and model performance was largely maintained (Fig.~\ref{fig:compare_scores} ). This result illustrates the distinction between model-specific and model-agnostic feature importance made in Part I.  3-5 km Max Refl is important to the original model due to its key compensating role, but for many other models in the Rashomon set, it does not have high importance relative to some of its correlated features (e.g., updraft speed, composite reflectivity). 

To objectively assess whether dimensionality reduction decreased the uncertainty in the top feature rankings, we introduce the weighted average feature ranking uncertainty ($\overline\sigma$):
\begin{equation}\label{eqn:rank_uncertainty}
    \overline{\sigma} = \frac{1}{\sum Median(R_j)} \sum_{j}^{P} \frac{\text{IQR}(R_j)}{\text{Median}(R_j)} 
\end{equation}
where $R_j$ is the set of ranks for feature $x_j$ ($|R| = 9$ for the WoFS dataset and $|R|$=10 for the road surface dataset), $P$ is the number of features, and IQR and Median are the interquartile range and median, respectively. By weighing by the median ranking, we ensure that rank uncertainty in the top features is weighed more than rank uncertainty in lower ranked features. To restrict our analysis, we only computed eqn \ref{eqn:rank_uncertainty} for the top 10 features based on the median rankings. 

The uncertainty in the rank of the top features was smaller for the reduced than the original models for all four datasets (Fig.~\ref{fig:median_rankings}). The rank uncertainty is expected to decrease for smaller feature sets as there are fewer possible ranks. However, other factors are likely contributing to the rank uncertainty reductions. For example, the severe hail dataset had the most significant drop in rank uncertainty (30.14 to 2.82), and this owes largely to improved rank certainty of the top 2 features. The decrease in ranking uncertainty was greater for the tornado and severe hail datasets (Fig.~\ref{fig:median_rankings}b,f) than for the severe wind and road surface datasets (Fig.~\ref{fig:median_rankings}d,h), perhaps due to the interaction strength reductions (Fig.~\ref{fig:compare_scores}b,d ). Reducing the strength of the feature interactions
enhances the contributions of first-order effects and thereby improving the estimate of feature-unique importance. 

\begin{figure*}[t]
    \noindent\includegraphics[width=38pc,angle=0]{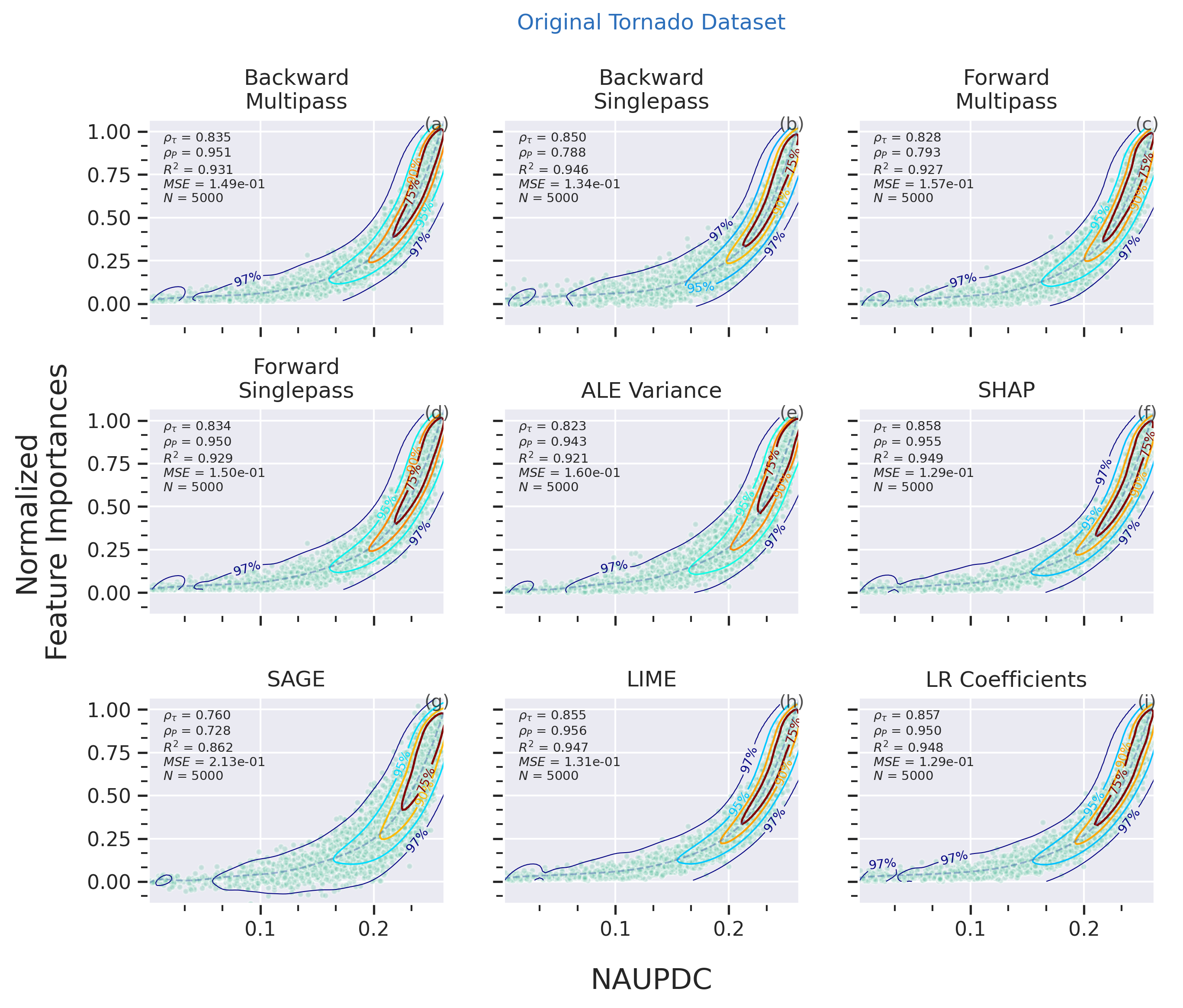}
    \caption{Relationship between the total importance\textemdash sum of feature importance scores for a given feature subset\textemdash and model performance of the re-trained models for the (a) backward multipass, (b) backward singlepass, (c) forward multipass, (d) forward singlepass, (e) variance of first-order ALE, (f) absolute sum of SHAP values, (g) SAGE, (h) LIME, and (i) the magnitude of the logistic regression coefficients. Total importance was scaled so that all scores ranged from 0 to 1. Results are valid for the original tornado dataset. To identify clustering, kernel density estimate contours are provided (e.g., the 97$\%$ contour contains 97$\%$ of the data). Listed in the upper left of each panel are the bootstrapped ($N$=100) mean Kendall rank correlation ($\rho_{\tau}$), linear correlation coefficient after log-transform ($\rho_P$), coefficient of determination ($R^2$), mean squared error (MSE) and the sample size $N$. MSE and $R^2$ are based on a 5th-degree polynomial fit.}
     \label{fig:experiment1_corr}
\end{figure*}

Given the large inter-method differences in the feature rankings highlighted in Part I, especially for the original datasets, we should question whether some feature ranking methods more faithfully assign importance than others (i.e., those features contributing strongly to the model prediction are assigned higher importance scores). Unfortunately, we do not know the true feature rankings so we cannot assess ground-truth faithfulness \citep{OpenXAI}. One alternative is to measure how feature importance scores correspond with model performance for a given dataset using the approach in \citet{Covert+etal2020}:
\begin{enumerate}
    \item Compute the feature importance/relevance score for each feature ranking method
    \item Randomly generate feature subsets (5000 in this study; subset size $\in$ [{1, $P$-1}] where $P$ is the number of features)
    \item For each feature subset, train a new model and evaluate the model performance (herein we use NAUPDC). 
    \item For each feature subset, compute the total importance score for each feature ranking method.

\end{enumerate}
One drawback to this approach is that the faithfulness of the feature rankings is not being fully measured with respect to the model-specific or model-agnostic feature importance. It is not model-specific as the model performances are based on the re-trained models, but it is not fully model-agnostic as the total importance scores are with respect to  the full datasets (either original or reduced feature sets). Nevertheless, this experiment does provide insight into whether higher importance is associated with higher model performance and lower importance with lower model performance. 

\begin{figure}[t]
    \noindent\includegraphics[width=20pc,angle=0]{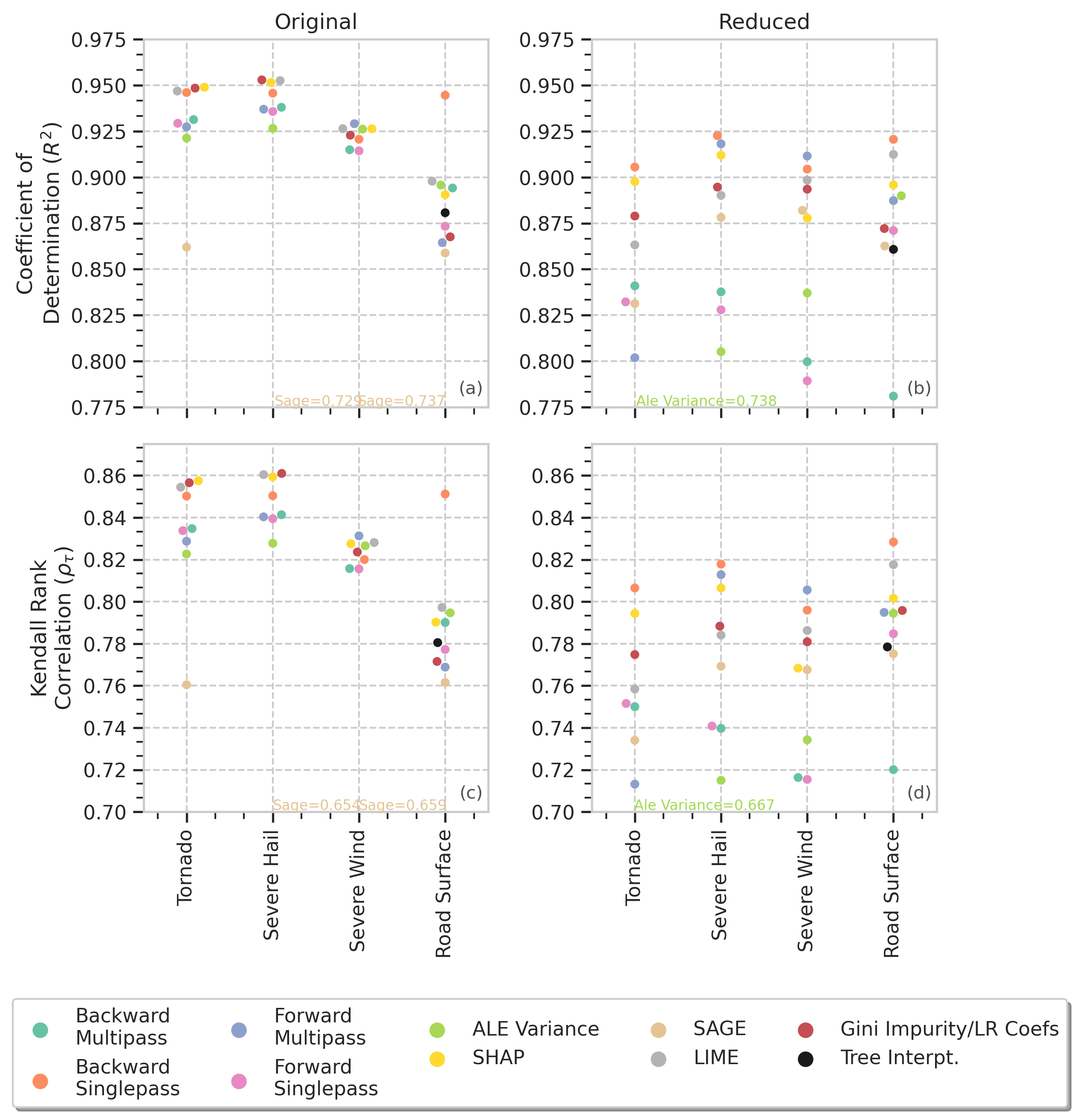}
    \caption{ Coefficient of determination ($R^2$; first row) and Kendall rank correlation ($\rho_{\tau}$) between model performance and total importance of re-trained models for the original (first column) and reduced (second column) datasets for each feature ranking method. To restrict the y-axis range, low values are set to y-axis minimum and annotated with their true value. To improve interpretation, the scatter points are adjusted slightly left or right of center to limit overlap. 
    }
     \label{fig:exp1}
\end{figure}

Fig.~\ref{fig:experiment1_corr} shows the normalized total importance against the model performance for each ranking method for the original tornado dataset (the results are fairly representative of the other datasets), while Fig.~\ref{fig:exp1} summarizes the results for each dataset. The total importance was normalized using maximum-minimum scaling to enable comparison amongst the methods. If a feature subset contains features with high importance, then we expect the model trained on that feature subset to have higher performance than an identically sized feature subset containing less-important features.

\citet{Covert+etal2020} only evaluated linear correlation between importance and model performance, but based on Fig.~\ref{fig:experiment1_corr}, we can see the relationship is highly non-linear and ranking faithfulness based on linear correlation would provide misleading results. Instead, we relied on multiple statistics to measure correspondence between the total importance and model performance: Kendall rank correlation coefficient ($\rho_{\tau}$), linear correlation after a log-transform ($\rho_{\rho}$), $R^2$, and mean squared error (MSE).  To compute the $R^2$ and MSE, we fit a 5-degree polynomial curve to the data and use goodness-of-fit ($R^2$; also known as the coefficient of determination) as a measure for feature ranking performance (in other words, we assume that the more the relationship between total importance and model performance matches a functional form, the better the method is at assigning individual feature importance). To summarize the feature ranking faithfulness, we limited our analysis to $R^2$ and $\rho_{\tau}$, as we felt they matched well with our subjective evaluation of the scatter plots for each dataset.

\begin{figure*}[t]
    \noindent\includegraphics[width=38pc,angle=0]{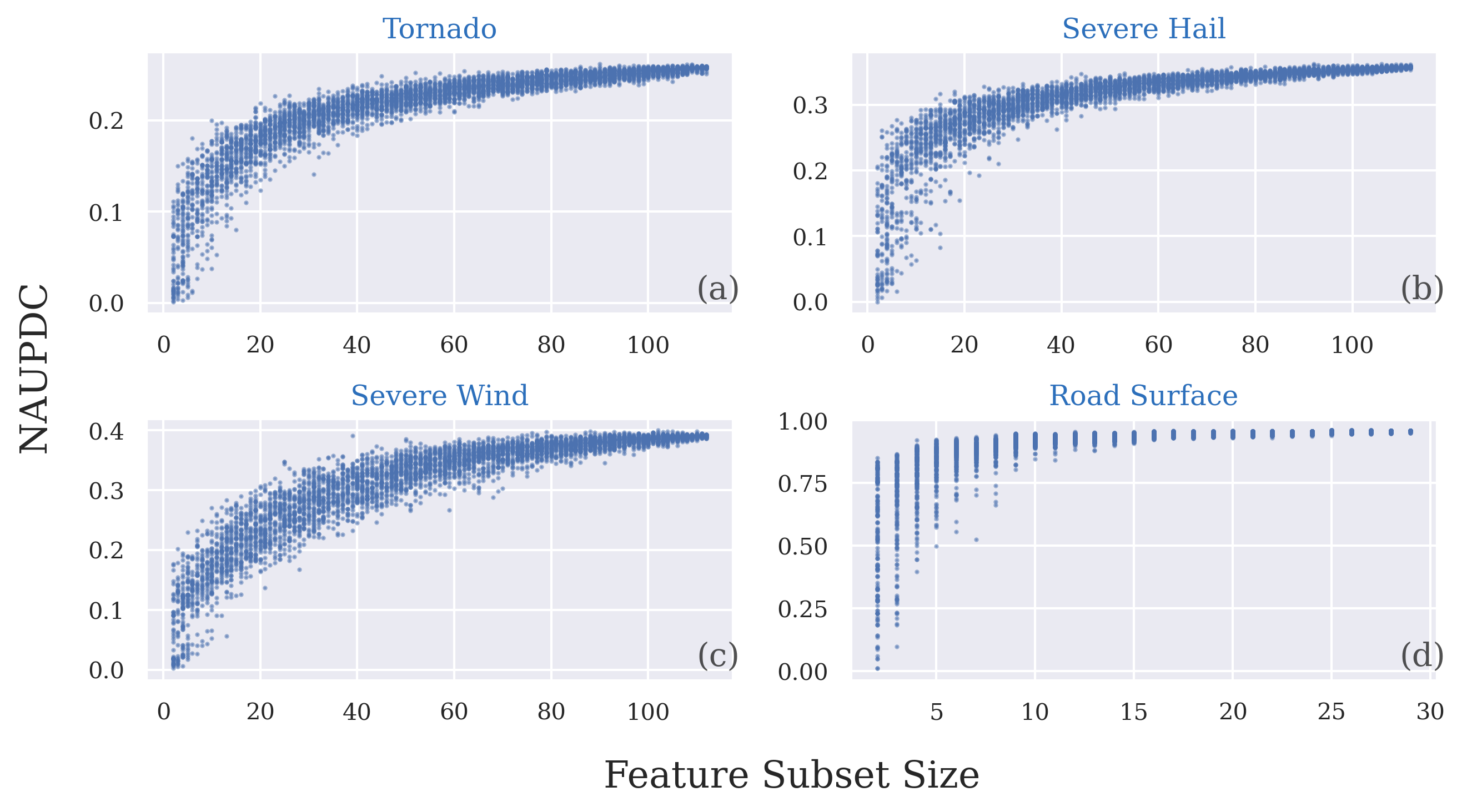}
    \caption{ Model performance (NAUPDC) versus feature subset size for the original (a) tornado, (b) severe hail, (c) severe wind, (d) road surface datasets. }
     \label{fig:pareto}
\end{figure*}

For the original datasets (Fig.~\ref{fig:exp1}a,c), there is a high correspondence between total importance and model performance for a majority of the methods and we find little separation between a majority of the feature rankings methods for both statistics. We know from Part I that the different feature ranking methods tend to have high agreement on the top features and the disagreement is largely based on exact ranking. These results suggest that those minor discrepancies in ranking do not amount to significant differences in the relationship between model performance and total importance. As for the strong correspondence between total importance and model performance, we can see the relationship is highly non-linear (Fig.~\ref{fig:experiment1_corr}) and most points are in the higher performance, higher importance region (e.g., NAUPDC > 0.2 and total importance > 0.5). In Fig.~\ref{fig:pareto}, we can see that model performance has an asymptotic relationship with feature subset size. The highly non-linear relationship suggests that the feature contributions to model performance follow a Pareto distribution \citep{DAVIS1979} such that a small subset of features are responsible for most of the model performance (Fig.~\ref{fig:pareto}). Thus, in a bulk sense, most of the feature ranking methods should be able to largely discriminate between strong and weak contributors to model performance. The Parteo-like distribution of feature importance also helps explain the success of the dimensionality reduction, as nearly 30 out of 113 features explain most of the model performance for the severe weather datasets (Fig.~\ref{fig:pareto}a-c) while approximately 10 explain the road surface dataset (Fig.~\ref{fig:pareto}d). 

In terms of specific methods, for all four original datasets, SHAP was one of the best method while SAGE scores had the least correspondence to model performance, despite both being based on Shapley theory (Fig.~\ref{fig:exp1}a,c). SHAP does not assume feature independence and leverages feature clustering based on correlations, which likely explains why it performs better than SAGE. Compared to the other feature ranking methods, forward single-pass (FSP) had a lower correspondence between total importance and model performance. Recall that the forward permutation methods starts with all features permuted, which breaks up the relationships between features. Neglecting those important relationships (e.g., potential compensating effects) and attempting to isolate individual importance appears to degrade the faithfulness of FSP. In terms of model-specific ranking methods, LR coefficients were among the top performers for the severe weather datasets while tree interpreter and Gini importance were poor performers for the road surface dataset. These results reinforce that the interpretability of logistic regression models can be leveraged to explain their behavior, but random forests are uninterpretable and require outside methods to understand and explain them. 

Despite the feature correlations, the backward single-pass (BSP) method was one of the most faithful, especially for the road surface dataset. This is not that surprising as the impact of correlated features on permutation importance scores is heavily dependent on the correlations between a feature and the target variable.
\citep{Gregorutti+etal2017_corr}. If two features are correlated, but one of them is a much stronger predictor of the target variable, then permutation importance scores may only be negligibly impacted by the feature correlations. To determine whether our datasets had sufficient correlations with the target variable compared to the correlations between features, we compare the average feature correlation (i.e., the expected correlation between any two features) and the average correlation of a feature with the target variable (Table ~\ref{tab:avg_corr}). For the road surface dataset, the average correlation with the target variable is similar to the average correlation with target variable while for the severe weather datasets, the features were more correlated with each than the target variable on average. The strong predictors of the target variable in the road surface dataset likely limited the negative impact of feature correlations, which helps explain the higher fidelity of BSP. For the severe weather datasets, the predictors are not nearly as strong as for the road surface dataset, but are likely sufficient to improve the fidelity of BSP for those datasets. 

\begin{table}[]
    \centering
    \begin{tabular}{lcc}
              \hline \hline 
              & Avg. Feature Correlations & Avg. Correlation with the Target  \\
              \hline 
         Hail & 0.22 & 0.12  \\
         Wind & 0.22 & 0.11 \\ 
         Tornado & 0.22 & 0.10 \\
         Road Surface & 0.20 & 0.19 \\
         \hline 
    \end{tabular}
    \caption{The average feature correlations and the average feature correlation with the target variable for all four datasets. }
    \label{tab:avg_corr}
\end{table}

\begin{figure*}[t]
    \noindent\includegraphics[width=38pc,angle=0]{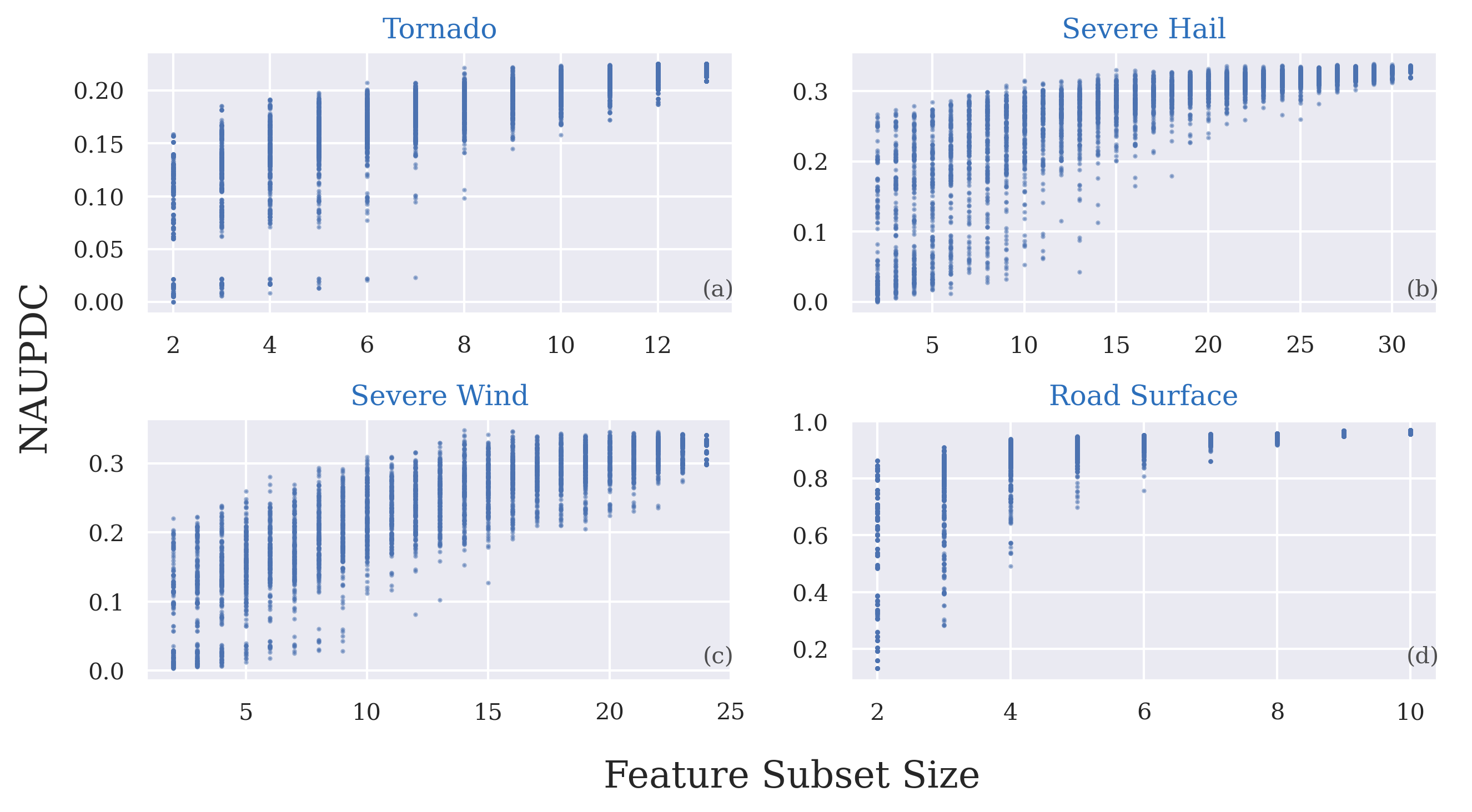}
    \caption{ Same as Fig.~\ref{fig:pareto}, but for the reduced feature sets. }
     \label{fig:pareto_reduced}
\end{figure*}

For the reduced datasets (Fig.~\ref{fig:exp1}b,d), $\rho_{\tau}$ and $R^2$ varied approximately between 0.65-0.9 and 0.75-0.92 (depending on the dataset), respectively, which are generally lower values and wider ranges than for the original datasets (cf. Fig.~\ref{fig:exp1}a,c and Fig.~\ref{fig:exp1}b,d). After dimensionality reduction, the feature importance scores were less Pareto-like (i.e., feature contributions to model performance were more even; Fig.~\ref{fig:pareto_reduced}). Thus, discriminating between low- and high-importance features is more difficult, consistent with the larger differences in performance between the methods. Though there is greater variability between the ranking methods, the LR coefficients, BSP, and SHAP are consistently among the most faithful (partially consistent with \citealt{Covert+etal2020}; did not include the LR coefficients in their study), while the ALE variance, FSP, and BMP are among the worst. The poor performance of FSP and ALE variance, even after reducing the feature sets and limiting feature correlations, suggests that isolating features (by permuting all other features or only including first-order effects, respectively) is a poor strategy for estimating feature importance. FMP is a top performer for the severe hail, wind, and road surface datasets, but amongst the worst for the tornado dataset. The improved faithfulness of the BSP, LR coefficients, and SAGE for the reduced feature sets is probably due at least in part to the reduced feature correlations. 

\begin{figure}[t]
    \noindent\includegraphics[width=38pc,angle=0]{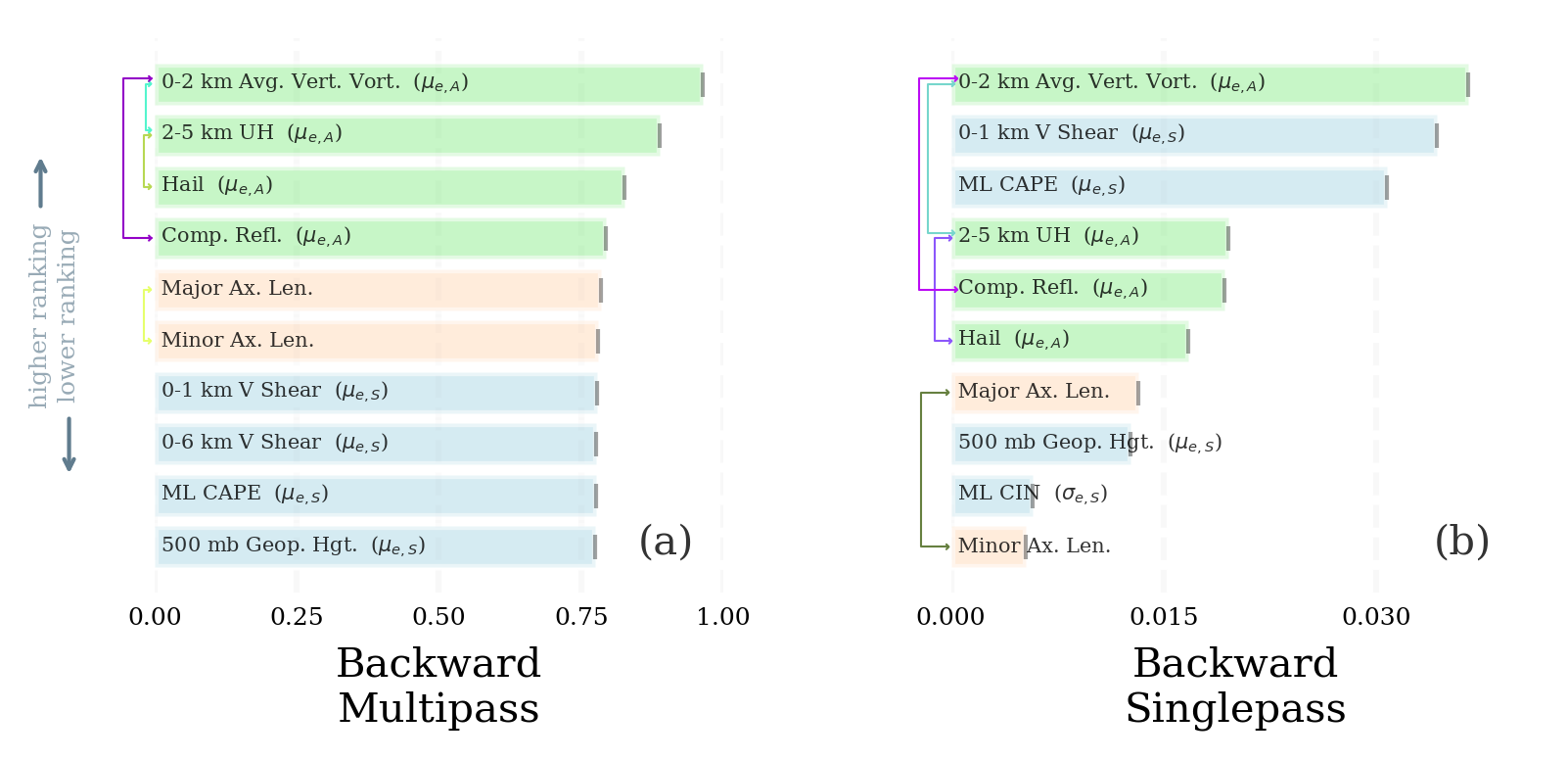}
    \caption{ Backward (a) multipass and (b) single-pass permutation importance ranking, respectively, for the reduced tornado dataset. Green, blue, and light gold indicate intra-storm, environment, and storm morphological features, respectively. Features with a linear correlation $>$ 0.5 are linked together. See Table 1 in \citet{Flora+etal2021} for the feature naming convention.}
     \label{fig:reduced_torn_rankings}
\end{figure}

Although the differences in $\rho_{\tau}$ and $R^2$ between the ranking methods may seem small, they correspond to substantial differences in the explanation of the model behavior. For example, for the reduced tornado dataset model, while both BSP and BMP (cf. Fig. ~\ref{fig:reduced_torn_rankings}a,b) have low-level vorticity as the most important feature (which is by default), the BSP ranks some of the intrastorm features substantially higher than the BMP does. Given BMP's lower fidelity implied by the above analysis, we should prefer the BSP rankings of these features, unless the BMP rankings are much more plausible (incidentally, in our judgment, they are not).

It is surprising that feature relevance methods tend to outperform or perform similarly to feature importance methods in assigning individual contributions to model performance, especially for the original datasets. The exceptions are Gini impurity and tree interpreter. For the Gini impurity, it is well established that the feature rankings are biased by the way the random forest is trained \citep{Strobl+etal2007, Strobl+etal2008}. Tree interpreter has a known bias of assigning lower attribution values for features higher up in the tree \citep{Lundberg+etal2020_localtoglobal}. We suspect the success of the SHAP owes to the Owen values accounting for co-dependencies amongst features, but more work is required to determine this. For the LR coefficients, we might expect instability as the dimensionality increases and the co-linear features become more prevalent. We hypothesize that the L1 and L2 regularizations mitigate this problem. The high fidelity of the LR coefficients for the original and reduced feature sets supports the use of partially interpretable models. 

\subsection{Assessing the Validity of Feature Rankings for Feature Selection.}
\begin{figure*}[t]
    \noindent\includegraphics[width=38pc,angle=0]{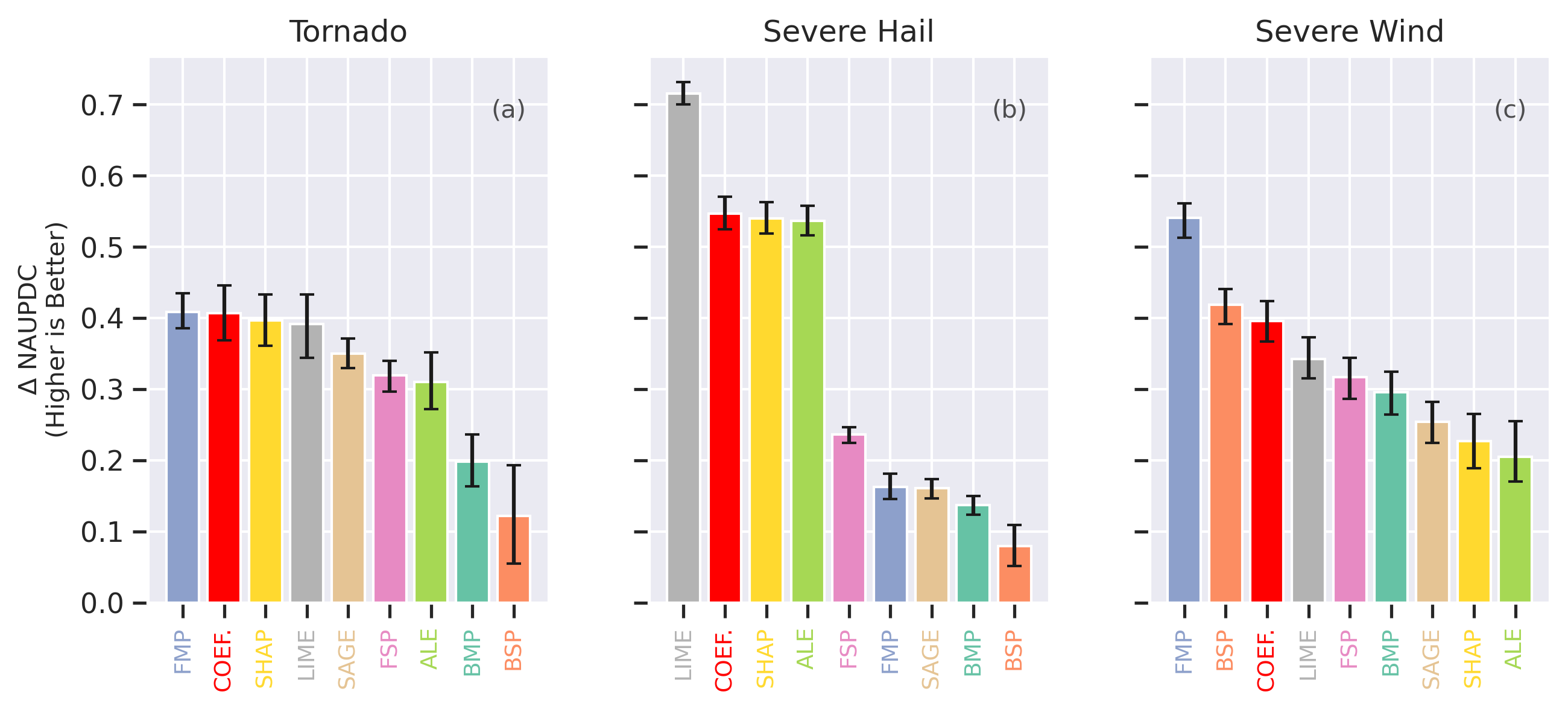}
    \caption{ Difference in NAUPDC between models trained on the top and bottom 15 features based on the original datasets. The bootstrapped ($N$=1000) 95$\%$ confidence intervals are shown in black. }
    \label{fig:exp2}
\end{figure*}

    The previous results highlight the overall performance of the ranking methods, but a more granular perspective is warranted. We now compare how well the methods discriminate between the top and bottom features by comparing the predictive power of the highest and lowest ranked features from each method. We train new models with the top and bottom 15 features, respectively, identified by each method for the original datasets and evaluate the models' performance on the training dataset (to be consistent with how feature importance is computed). Given the strong agreement among the methods for the original road surface temperature dataset (Fig. ~\ref{fig:median_rankings}g; Fig. 4 from Part I) and that the dataset only has 30 features, we restrict our focus to the WoFS-based datasets. For severe hail and wind, the method that produced the highest correlation between total importance and model performance (Fig.~\ref{fig:exp1}a,c) also produced the greatest difference in performance between the highest and lowest ranked features (Fig.~\ref{fig:exp2}b,c), but the best method was hazard dependent (LIME for hail and FMP for wind). For the tornado dataset, the most faithful feature ranking methods also discriminated well between the best and worst features, but so did FMP, which had relatively low fidelity (Fig.~\ref{fig:exp1}c). Another surprising result is that SHAP discriminated poorly between the best and worst features  (Fig.~\ref{fig:exp2}c) despite being a top overall performer for the severe wind dataset (Fig.~\ref{fig:exp1}g). 
    
    \begin{figure*}[t]
    \noindent\includegraphics[width=38pc,angle=0]{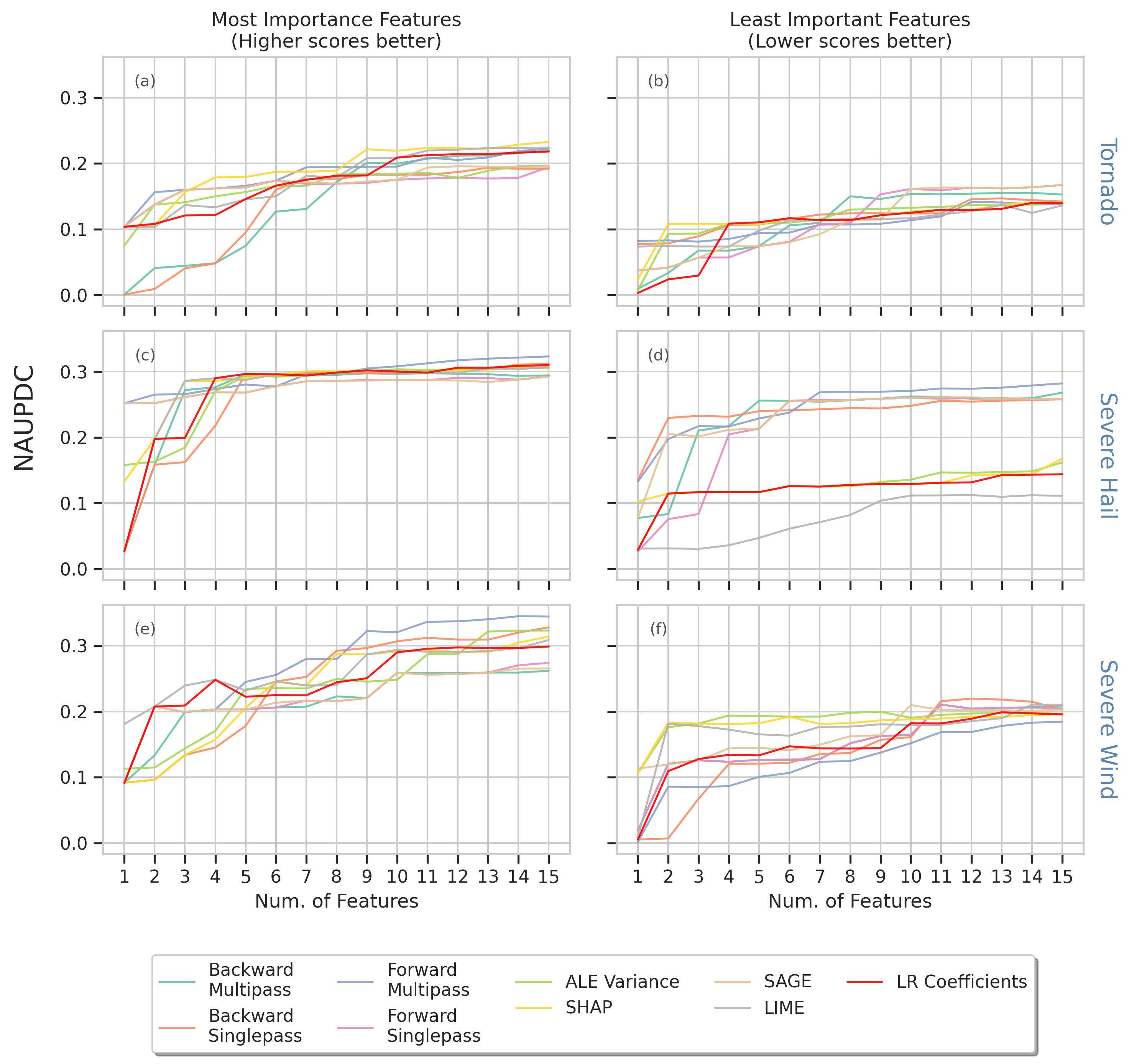}
    \caption{ Performance of feature selection for the most important features up to the top 15 (left column) and (right column) the least important features up to the worst 15 for the tornado (top row), severe hail (middle row), and severe wind (bottom row) for each feature ranking method.  }
    \label{fig:feature_selection}
    \end{figure*}
    
    To better appreciate the results in Fig.~\ref{fig:exp2}, we retrain a model on the top feature, top 2 features, and so on until we reach the top 15 and compute the model performance (we repeat this for the worst feature, worst 2 features, and so on until we reach the worst 15).  From a feature selection perspective, FMP performs well in determining the features that contribute the most to the model skill for the tornado dataset (Fig.~\ref{fig:feature_selection}a), but performs poorly at determining the features that contributed the least to the model performance (Fig.~\ref{fig:feature_selection}b). As for BSP and BMP, one could argue that their top features are important largely due to their association with the remaining features (more so for BMP; \citealt{Lakshmanan+etal2015}) and thus would not perform well as feature selection methods. Our analysis supports this notion: BSP and BMP have a less faithful set of top five predictors compared to the other methods, although for the top 10+ features they are more similar (Fig. ~\ref{fig:feature_selection}a). When examining the relationship between the top BSP/BMP features (not shown) and tornado likelihood, we found that the top BSP/BMP features have poorer discrimination between tornado and non-tornado (not shown) compared to top features from the other methods. We suspect that other, more important features are lower ranked due to their correlations with other features, a well-known issue with backward permutation importance \citep{Gregorutti+etal2015_grouped, Strobl+etal2007, Molnar+etal2020_imlpaper, InterpretMLTextbook}. 

    For severe hail (Fig.~\ref{fig:feature_selection}c,d), LIME had the greatest performance difference between the top and worst 15 features, which is due to LIME better identifying the features that contribute the least to model performance (Fig.~\ref{fig:feature_selection}d). In fact, the feature relevance methods (LIME, ALE variance, SHAP, and LR coefficients) all performed better than the feature importance methods in identifying the least important features. However, feature relevance methods such as SHAP and ALE variance underperformed feature importance methods in identifying the least important features in the severe wind dataset (Fig.~\ref{fig:feature_selection}f). 
    
    The two main takeaways from this section are that (1) the general fidelity of a feature ranking vary for more granular explanations (e.g., FMP had poor bulk fidelity for the original tornado dataset, but discriminated well between the top and worst 15 features) and (2) feature importance/relevance methods can perform well as feature selection methods, but the behavior is inconsistent. This is not a surprising result, as many of these methods are designed to measure model-specific feature importance, which may not translate to model-agnostic feature importance. 

\subsection{Can we reduce feature ranking uncertainty by leveraging the relative fidelity of the feature ranking methods? }
Figs.~\ref{fig:exp1} and ~\ref{fig:exp2} show that there are disparities between feature rank and contribution to model performance for the different ranking methods. Thus, we determine whether the uncertainty of the feature rank is meaningfully reduced by excluding less faithful ranking methods. To assess the improvement in rank uncertainty, we introduce the ranking uncertainty ratio:
\begin{equation}\label{eqn:ratio}
    \text{Ranking Uncertainty Ratio} = \frac{\overline{\sigma_{top}}}{ \displaystyle \mathop{\mathbb{E}}(\overline{\sigma_{subsets}})} 
\end{equation}
The numerator is the feature rank uncertainty (eqn. \ref{eqn:rank_uncertainty}) re-computed per dataset using the top 3 ranking methods based on Fig.~\ref{fig:exp1}, while the denominator is the expected rank uncertainty from all combinations of 3 methods from among all our ranking methods. Ideally, the ranking uncertainty ratio should be $< 1$. 

\begin{figure}[t]
    \noindent\includegraphics[width=20pc,angle=0]{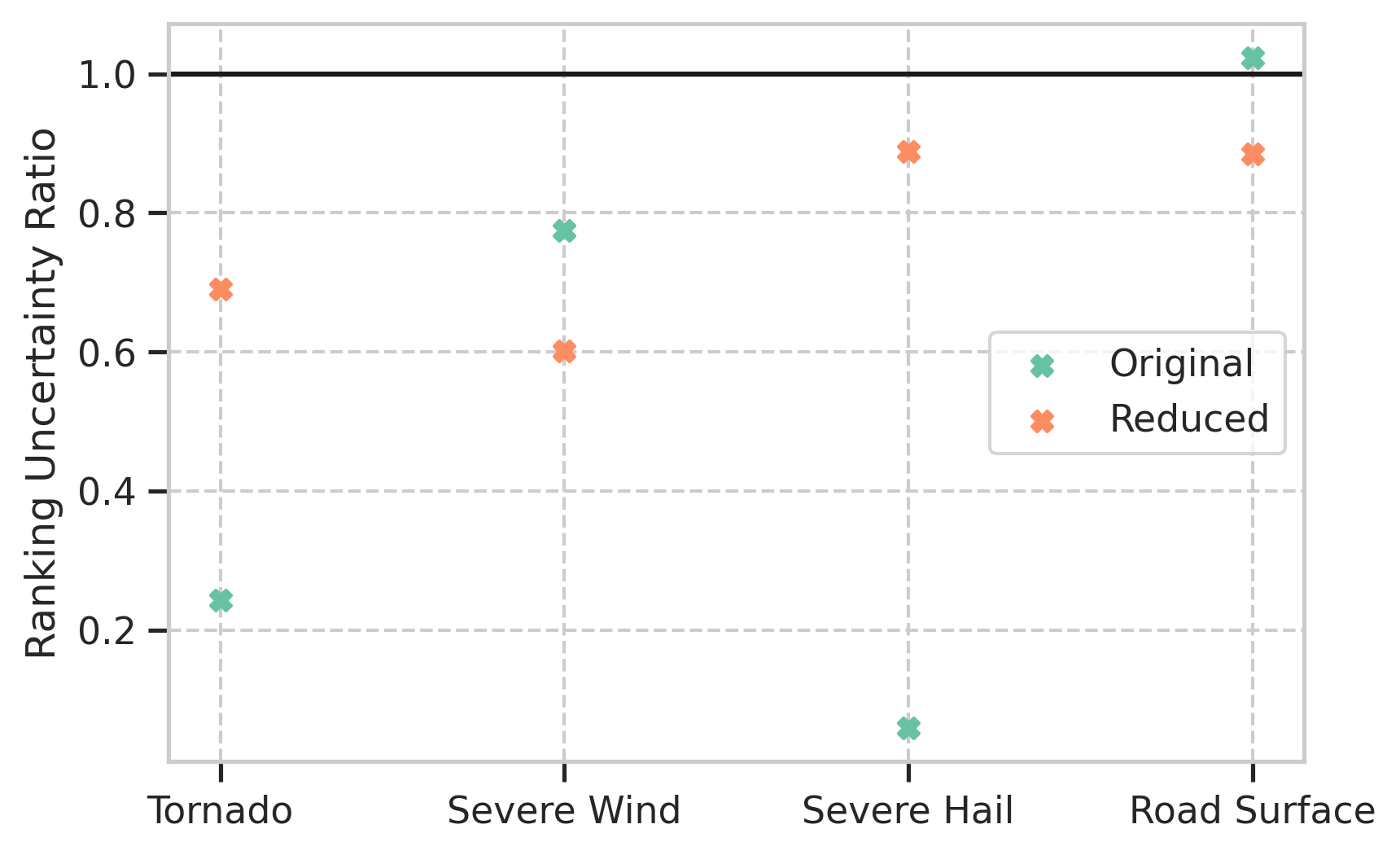}
    \caption{ The ranking uncertainty ratio (eqn \ref{eqn:ratio}) for the original (green) and reduced datasets (orange). A ranking uncertainty ratio of 1 is highlighted with a black line.}
     \label{fig:ratios}
\end{figure}

For the original severe wind and road surface datasets, the ratio is close to 1 while for the original severe hail and tornado datasets the ratio is 0.20 and 0.68, respectively (Fig.~\ref{fig:ratios}). It is not surprising that the rank uncertainty ratio for the road surface dataset is approximately 1 (1.02) as the general rank uncertainty in the original dataset was fairly low (Fig.~\ref{fig:median_rankings}) and the methods were largely in agreement on the top features (Part I). The low rank uncertainty ratios for the original tornado and severe hail datasets suggest that the uncertainty in the feature rankings of the original tornado and severe hail datasets was likely inflated by including less faithful ranking methods. 

The ranking uncertainty ratios are $<$ 1 for the reduced datasets. It is not surprising that the ratios for the reduced tornado and severe hail datasets are higher than for the original datasets since the rank uncertainty across all methods had already been substantially decreased by dimensionality and interaction strength reduction (Fig.~\ref{fig:compare_scores} and Fig. ~\ref{fig:median_rankings}). These higher ratios can also be partially explained by the fact that for the reduced feature datasets, the set of possible rankings is substantially reduced, and thus the expected uncertainty (the denominator in eq. \ref{eqn:ratio}) is smaller. Nevertheless, these results suggest that knowing the relative faithfulness of feature ranking methods can reduce feature rank uncertainty. 

\section{Conclusions}\label{sec:conclusions}
ML models are becoming increasingly common in the atmospheric science community with a wide range of applications. ML models are powerful statistical tools, but their complexity often prevents a comprehensive understanding of their internal functionality. Improving model understanding is necessary, as decision makers will need to interrogate these models to begin to establish trust in them, especially in high-risk situations where decisions can be costly. In Part I, we highlighted several post hoc explainability methods (e.g., SHAP, LIME, permutation importance, partial dependence/ALE) that have been developed to improve our understanding of complex ML models.

Little has been done, however, to verify the faithfulness of the explainability methods. In this Part II of a two-part study, we evaluated and compared multiple feature ranking methods using a convection-allowing ensemble (WoFS) severe weather dataset and a nowcasting road surface dataset derived from the HRRR model. Given the sensitivity of many of these methods to correlated features, we also evaluated how the feature ranking faithfulness is impacted by dimensionality reduction. This is one of the first studies to quantify how model explainability can improve through dimensionality reduction and knowing the relative faithfulness of feature ranking methods for a given dataset.

The conclusions are as follows: 
\begin{itemize}

    \item For all datasets, dimensionality reduction objectively reduced model complexity (i.e., reduced feature interaction strength and first-order effects; \citealt{Molnar+etal2019}) without greatly impacting model performance. This is a promising result as increasing explainability is becoming a key issue for model development and ethical use of AI. 

    \item Feature rankings varied considerably among the explainability methods, but the various rankings were all plausible. Thus, confirmation bias could mislead domain experts into trusting the feature rankings assigned by individual methods. 
    
    \item The average feature ranking uncertainty (i.e., the variation in rank averaged for all features) was substantially decreased in the reduced versus the original datasets. For the severe hail dataset, the uncertainty was reduced by a factor of 10 while the other datasets had reductions of 2-3. This is one of the first attempts to quantify how dimensionality reduction improves explainability. 
    
    \item The average feature ranking uncertainty was also reduced by excluding lower fidelity ranking methods. This result suggests that knowing the relative faithfulness of the explainability methods for a given dataset can lead to improved explanations.
    
    \item The traditional permutation importance (backward singlepass) was a reliable method for all datasets, even without dimensionality reduction despite prevalent feature correlations. As discussed in Part I, the impact of correlations between features can be mitigated if the features are strong predictors of the target variable. 
    
    \item Top performers for both the original and reduced severe weather datasets include SHAP and LR coefficients. The faithfulness of LR coefficients motivates the use of partially interpretable models since they can be partially understood without the use of post-hoc explainability methods. The same cannot be said for model-specific explainability (e.g., tree-based methods) as the Gini impurity and tree interpreter methods were among the least faithful methods for both the original and reduced road surface datasets. 
    
    \item For the datasets and models used in this study, the feature relevance methods (SHAP, LR coefficients) tended to outperform or perform similar to the feature importance methods. The forward singlepass and backward multipass were consistently amongst the worst ranking methods (often failing to improve upon the backward singlepass method) while the other flavors of permutation importance were found to have inconsistent behavior, possibly depending on whether the dataset/model was largely based on first-order effects. 

    \item The differences in fidelity between the feature ranking methods became smaller as the dimensionality increased, which is inconsistent with \citet{Covert+etal2020}. Since our datasets follow a Pareto distribution (i.e., a small subset of features are responsible for most of the model performance), we expect the differences in fidelity between feature rankings methods to be smaller as the dimensionality increases. This highlights that when dealing with large feature sets, a more granular approach to exploring model explainability fidelity is required. 
    
    \item It is possible to leverage feature ranking methods as feature selection methods. This is a promising result as other methods of robust feature selection are often computationally prohibitive. 
    
\end{itemize}

This study is one of the first to explore the faithfulness of feature importance methods using geoscience-based datasets, and there are several promising avenues for future work.  First, we only applied these techniques to two datasets and two different classes of ML models (i.e., logistic regression and random forests). It is unclear how well the results extend to other datasets and models (e.g., deep learning models). Second, it will be necessary to collaborate with social scientists and end users (e.g., forecasters) to determine how to balance the tradeoff between explainability and skill. Complex models should only be favored if the performance gain is practically significant \textemdash a judgment that the practitioner must make. Third, though we demonstrated a method introduced by \citet{Covert+etal2020} for exploring the fidelity of explainability methods when the ground truth is unknown, it would be useful to assess the skill of explainability methods on benchmark datasets where the ground truth is known \citep{Mamalakis2021, OpenXAI}. 

\clearpage
\acknowledgments
Funding was provided by NOAA/Office of Oceanic and Atmospheric Research under NOAA-University of Oklahoma Cooperative Agreement $\#$NA21OAR4320204, U.S. Department of Commerce. The authors thank Eric Loken for informally reviewing an early version of the manuscript. We also acknowledge the team members responsible for generating the experimental WoFS output, which include Kent Knopfmeier, Brian Matilla, Thomas Jones, Patrick Skinner, Brett Roberts, and Nusrat Yussouf.  This material is also based upon work supported by the National Science Foundation under Grant No. ICER-2019758. Any opinions, findings, and conclusions or recommendations expressed in this material are those of the author(s) and do not necessarily reflect the views of the National Science Foundation. 

\datastatement
The experimental WoFS ensemble forecast data and road surface dataset used in this study are not currently available in a publicly accessible repository. However, the data and code used to generate the results here are available upon request from the authors. The explainability methods were computed and visualized using the scikit-explain python package \citep{Flora+Handler} developed by Dr. Montgomery Flora and Shawn Handler. The python scripts used to generate the figures in Part I and II are available at https://github.com/monte-flora/compare-explain-methods. 
\appendix 
\appendixtitle{Hyperparameter Optimization}

Hyperparameter selection was performed using the Bayesian hyperparameter optimization available in the hyperopt Python package \citep{Bergstra2013}. Details for hyperparameter optimization are provided in \citet{Flora+etal2021}. The hyperparameters for the models trained on the reduced feature sets are presented in Table \ref{table:parameters_first_hour}. 

\begin{table}[h]
\caption{Hyperparameter values used for the reduced feature set models in this study. The hyperparameter values for the original models are provided in \citet{Flora+etal2021} and \citet{Handler+etal2020}. The hyperparameter $C$ controls the overall regularization strength (smaller values indicate stronger regularization) and $\rho$ dictates if the regularization is largely L1 or L2 (values between 0-1; closer to 0 indicate stronger L2 regularization).  } \label{table:parameters_first_hour}
\begin{center}
\begin{tabular}{llccc}
\hline \hline
                    & Hyperparameter & Tornadoes & Severe hail & Severe Wind  \\
\hline                 
              & C & 0.1 & 0.01 & 0.01  \\
              & $\rho$ (l1$\_$ratio) & 0.0001 & 0.01 & 0.001   \\ 

\hline
            &  & Road Surface &  &  \\
            & Num. of Trees &  500             &  &   \\
            & Depth         & 20              &  &   \\
            & Max Features   & 5            &  &   \\
            & Min Sample Leaf & 5              &  &   \\
            & Min. Sample Split &   8       &  &   \\
            & Criterion      &   entropy       &  &   \\
            & Class Weight    &   Balanced       &  &   \\
\hline
\end{tabular}
\end{center}
\end{table}

\bibliographystyle{ametsocV6}
\bibliography{references}

\end{document}